\pdfoutput=1

\documentclass[11pt]{article}

\usepackage{acl}

\usepackage{times}
\usepackage{latexsym}

\usepackage[T1]{fontenc}

\usepackage[utf8]{inputenc}

\usepackage{microtype}

\usepackage{multirow}
\usepackage{multicol}

\usepackage{graphicx}
\usepackage{bm}
\usepackage{bbm}
\usepackage{amsmath}
\usepackage{amsfonts}
\usepackage{booktabs}
\usepackage{tikz}
\usepackage{makecell}
\usepackage{float}
\usepackage{subfigure}
\usepackage{color,xcolor}
\usepackage{amssymb}
\usepackage{utfsym}
\usepackage{enumitem}

\usepackage{colortbl}
\definecolor{ggray}{HTML}{E7E6E6}
\definecolor{ggreen}{HTML}{d4e7cf}

\title{LCS: A Language Converter Strategy for Zero-Shot Neural Machine Translation}

\author{
  Zengkui Sun\textsuperscript{1}\thanks{ \ \ Work was done when Zengkui Sun was an intern at Pattern Recognition Center, WeChat AI, Tencent Inc, China.},
  Yijin Liu\textsuperscript{2},
  Fandong Meng\textsuperscript{2},
  Jinan Xu\textsuperscript{1},
  Yufeng Chen\textsuperscript{1}\thanks{ \ \ Yufeng Chen is the corresponding author.}
  and Jie Zhou\textsuperscript{2} \\
  \textsuperscript{1}Beijing Jiaotong University, China \\
  \textsuperscript{2}Pattern Recognition Center, WeChat AI, Tencent Inc, China \\
  \texttt{\{zengksun,jaxu,chenyf\}@bjtu.edu.cn} \\
  \texttt{\{yijinliu,fandongmeng,withtomzhou\}@tencent.com} \\
}

\begin{document}
\maketitle
\begin{abstract}
Multilingual neural machine translation models generally distinguish translation directions by the language tag (LT) in front of the source or target sentences.
However, current LT strategies cannot indicate the desired target language as expected on zero-shot translation, \emph{i.e.}, the \textit{off-target} issue.
Our analysis reveals that the indication of the target language is sensitive to the placement of the target LT.
For example, when placing the target LT on the decoder side, the indication would rapidly degrade along with decoding steps, while placing the target LT on the encoder side would lead to copying or paraphrasing the source input.
%
To address the above issues, we propose a simple yet effective strategy named \textbf{L}anguage \textbf{C}onverter \textbf{S}trategy (\textbf{LCS}).
By introducing the target language embedding into the top encoder layers, LCS mitigates confusion in the encoder and ensures stable language indication for the decoder.
Experimental results on MultiUN, TED, and OPUS-100 datasets demonstrate that LCS could significantly mitigate the \textit{off-target} issue, with language accuracy up to 95.28\%, 96.21\%, and 85.35\% meanwhile outperforming the vanilla LT strategy by 3.07, 3,3, and 7.93 BLEU scores on zero-shot translation, respectively.
\end{abstract}

\section{Introduction}
Multilingual Neural Machine Translation (MNMT) aims to build a unified model to support the translation between any language pairs \citep{dabre2020survey, zhang2021competence, fan2021beyond}.
One main challenge is the indication of the translation direction.
Pioneers \citep{dong-etal-2015-multi, luong-etal-2015-effective, firat-etal-2016-multi, lu-etal-2018-neural} utilize the language-specific encoder or decoder to distinguish source or target language.
To simplify the architecture, \citet{firat-etal-2016-zero} and \citet{johnson-etal-2017-googles} propose the language tag (LT) strategy, which places an artificial language tag in front of the source input sentence, to indicate the desired target language without modification on the vanilla NMT architecture and training objective.
Due to its simplicity and efficiency, the LT strategy has become the \textit{de facto} strategy to build unified MNMT models \citep{johnson-etal-2017-googles, dabre2020survey, zhang-etal-2020-improving, fan2021beyond}, even other unified models \citep{zhang-etal-2023-bridging, wang2022clidsum, liang2023unified}. 

\begin{table}[t]
\centering
\resizebox{0.48\textwidth}{!}{
    \begin{tabular}{c | l l }
    \toprule
    \textbf{LT Strategies} & \multicolumn{2}{c}{\textbf{Translation Pair (\textcolor{blue}{De}$\to$\textcolor[rgb]{0.1,0.7,0.2}{Fr})}} \\
    \midrule
    \textit{Source}  & Wie sind sie durch die Sicherheitskontrolle gekommen? & \textcolor{blue}{\textit{(German)}} \\
    \textit{Reference} & Comment avez-vous passé le contrôle de sécurité? & \textcolor[rgb]{0.1,0.7,0.2}{\textit{(French)}}\\
    \midrule
    T-Enc & Wie sind sie durch die Sicherheitskontrolle gekommen? & \textcolor{blue}{\textit{(German)}}\\
    \midrule
    S-Enc-T-Dec & How'd they get through the security? & \textcolor{red}{\textit{(English)}}\\
    \midrule
    LCS (ours) & Comment sont-elles venues par les contrôles de sécurité? & \textcolor[rgb]{0.1,0.7,0.2}{\textit{(French)}} \\
    \bottomrule
    \end{tabular}
}
\caption{
    Translation results of the same translation pair (\textcolor{blue}{German}$\to$\textcolor[rgb]{0.1,0.7,0.2}{French}) with three LT strategies\textsuperscript{\ref{footnote: lt}}. T-Enc leads to mistranslating into the undesired \textcolor{blue}{German} sentence and S-Enc-T-Dec leads to mistranslating into the \textcolor{red}{English} sentence, while LCS accurately leads to translating into the \textcolor[rgb]{0.1,0.7,0.2}{French} sentence.
}
\label{tab: case_study}
\end{table}

With the LT strategy, the MNMT models theoretically support many-to-many translation, and even zero-shot translation \cite{johnson-etal-2017-googles, dabre2020survey, gao2023improving}.
However, in practice, the MNMT models frequently mistranslate the source language to the wrong target language on zero-shot translation, referred to as the \textit{off-target} issue \citep{zhang-etal-2020-improving}.
Tab.\ref{tab: case_study} shows an example of zero-shot translation from German to French, in which T-Enc and S-Enc-T-Dec\footnote{S\,/\,T represents the source\,/\,target LT, and Enc\,/\,Dec represents the LT placed on the Encoder\,/\,Decoder side. Tab.\ref{tab: lt_strategy} shows more detailed examples.\label{footnote: lt}} are commonly used LT strategies \citep{johnson-etal-2017-googles, fan2021beyond, wu-etal-2021-language}. 
In this case, neither of both strategies could help MNMT models translate the sentence into the correct target language.
The T-Enc strategy leads to the \textit{To-Source} issue (\emph{i.e.}, paraphrase or copy the input of source language), while S-Enc-T-Dec leads to the \textit{To-English} issue (\emph{i.e.}, translate into English).
For the \textit{To-Source} issue, prior studies \citep{wang2022understanding} suggest that the data noises make this issue.
However, our experiments display that the \textit{To-Source} issue still exists after data denoise and the different language tag strategies result in different error distribution on both issues (\S\ref{sec: off_target}).
And for the \textit{To-English} issue, we suspect it comes from the inadequate language indication of the target language and the most-common language (\emph{i.e.}, English) is model preferred.

To further investigate the above issues, we conduct experiments to explore the indication of the target language in various LT strategies.
And we summarize some preliminary findings:
\begin{itemize}[leftmargin=*,topsep=0pt]
\setlength{\itemsep}{0pt}
\setlength{\parsep}{0pt}
\setlength{\parskip}{0pt}
    \item Placing the target language tag on the encoder side yields a more stable indication, while placing it on the decoder side delivers a decreasing indication throughout decoding steps and results in the \textit{To-English} issue (\S\ref{sec: fine_graind_acc}).
    \item The encoder tends to convert the states to be target language-specific on the top layers, and placing the target LT at the top encoder layers could mitigate the \textit{To-Source} issue (\S\ref{sec: variation}).
    \item Unfortunately, mainstream LT strategies suffer at least one issue between the \textit{To-Source} and \textit{To-English} in different degrees, even data denoising or placing the target language tag at the top encoder layers (\S\ref{sec: off_target}\,\&\,\S\ref{sec: variation}).
\end{itemize}

On these grounds, we propose a simple yet effective strategy, \textbf{L}anguage \textbf{C}onverter \textbf{S}trategy (\textbf{LCS}), to address the above issues.
Specifically, we first split the encoder layers and introduce the target language information to the deeper layers, which is named Language Converter by us.
Compared to the mere placement of language tag, we supply the target language embedding, which contains language-specific features~\citep{bjerva-etal-2019-language, oncevay-etal-2020-bridging, jin2022informative}, into each input state in each layer of the Language Converter layers.
%
In this manner, LCS could provide stable and sufficient target language indication for MNMT model, avoiding both \textit{To-Source} and \textit{To-English} issues.

Experimentally, LCS could significantly mitigate the \textit{off-target} issue and boost the performance of zero-shot translation.
Specifically, compared to the most widely-used strategy, \emph{i.e.}, the T-Enc strategy, LCS effectively mitigates the \textit{off-target} issue by improving language accuracy up to 95.28\% (\textit{+2.7\%}), 96.21\% (\textit{+1.71\%}), 85.35\% (\textit{+40.97\%}), and 86.67\% (\textit{+5.03\%}) on zero-shot translation of MultiUN, TED, OPUS-100 (noise and denoised) datasets, respectively.
Furthermore, LCS outperforms the T-Enc strategy by 3.07, 3.3, 7.93, and 2.93 BLEU scores improvements on zero-shot translation of these datasets, respectively. 
Meanwhile, LCS maintains the performance of the supervised translation and performs well on the noise data.
Moreover, LCS is well compatible with other approaches, \emph{e.g.}, denoising-encoder~\citep{wang-etal-2021-rethinking-zero}, contrastive learning~\citep{pan-etal-2021-contrastive}, LEE~\citep{jin2022informative}, and mBART~~\cite{liu-etal-2020-multilingual-denoising}, and yields further improvements when applied to them.

The main contributions of this paper can be summarized as follows\footnote{Codes are released at \url{https://github.com/Acerkoo/LCS}.}:
\begin{itemize}[leftmargin=*,topsep=0pt]
\setlength{\itemsep}{0pt}
\setlength{\parsep}{0pt}
\setlength{\parskip}{0pt}
    \item We take the analysis of the \textit{To-Source} and \textit{To-English} issues in the \textit{off-target} issue, and explore their causes in terms of the language representation variation of the encoder and language generation.
    \item We propose a simple yet effective strategy, LCS, to address the \textit{off-target} issue and further improve zero-shot translation quality, without introducing extra parameters.
    \item Experimental results demonstrate that LCS could significantly mitigate the \textit{off-target} issue and further boost the performance of zero-shot translation, with strong compatibility.
\end{itemize}

\section{Background}
\subsection{Multilingual Neural Machine Translation}
In bilingual neural machine translation (NMT), given a source sentence with $n$ tokens $\mathbf{x} = \{x_1, x_2, \dots, x_n\}$ and its target sentence with $m$ tokens $\mathbf{y} = \{y_1, y_2, \dots, y_m\}$,
and NMT models are generally optimized by the cross-entropy loss:
\begin{equation}\label{eq:ce_loss}
    \mathcal{L}_{\rm NMT} (\theta) = -\sum_{j=1}^{m} \log p(y_j \vert \mathbf{y}_{< j}, \mathbf{x}; \theta),
\end{equation} 
where $j$ is the index of each decoding step, $\mathbf{y}_{<j}$ is the target-side previous context for $y_j$, and $\theta$ represents the model parameter.

While in MNMT, for the language pair $(\mathbf{x}^s$, $\mathbf{y}^t)$ where $s$ and $t$ represent the source language $s$ and the target language $t$, and $\mathbf{t}$ denotes the target language tag, the MNMT models are generally optimized by the cross-entropy loss:
\begin{equation}\label{eq:m_ce_loss}
    \mathcal{L}_{\rm MNMT} (\theta) = -\sum_{j=1}^{m} \log p(y_j \vert \mathbf{y}^{t}_{< j}, \mathbf{x}^s, \mathbf{t}; \theta).
\end{equation}

In this paper, we focus on the analysis of the placement of $\mathbf{t}$ and explore the better way to indicate the target language for zero-shot translation.

\begin{table}[t!]
\centering
\resizebox{0.475\textwidth}{!}{
    \begin{tabular}{c | l }
    \toprule
    \textbf{LT Strategies} & \textbf{\makecell[c]{Example (En $\to$ De)}} \\
    \midrule
    \multirow{2}*{\makecell[c]{T-Enc\\\citep{johnson-etal-2017-googles}}} & <de> Hello, how are you? \\
    ~ & Hallo, wie geht's? \\
    \midrule
    \multirow{2}*{\makecell[c]{S-Enc-T-Dec\\\citep{fan2021beyond}}} & <en> Hello, how are you? \\
    ~ & <de> Hallo, wie geht's? \\
    \midrule
    \multirow{2}*{\makecell[c]{ST-Enc\\\citep{xue-etal-2021-mt5}}} & <en> <de> Hello, how are you? \\
    ~ & Hallo, wie geht's? \\
    \midrule
    \multirow{2}*{\makecell[c]{ST-Enc-T-Dec\\\citep{elnokrashy2022language}}} & <en> <de> Hello, how are you? \\
    ~ & <de> Hallo, wie geht's? \\
    \bottomrule
    \end{tabular}
}
\caption{
    Several examples of language tag strategies.
}
\label{tab: lt_strategy}
\end{table}
\subsection{Language Tag Strategy} \label{sec: lt_strategy}
Pioneered by \citet{johnson-etal-2017-googles}, the LT strategy has become the \textit{de facto} strategy to build the unified MNMT model.
Recently, many varieties are proposed to adjust the placement of the LT and we list several popular samples in Tab.\ref{tab: lt_strategy}.
Among these strategies, the T-Enc and S-Enc-T-Dec strategies are the most widely used ones to build MNMT models.
The first proposed T-Enc strategy performs the best on zero-shot translation \citep{wu-etal-2021-language, wicks2022effects}.
And S-Enc-T-Dec is also widely used in establishing MNMT models, \emph{e.g.}, M2M100 \citep{fan2021beyond}, mBART \citep{liu-etal-2020-multilingual-denoising}, and so on.
Although some studies \citep{xue-etal-2021-mt5, elnokrashy2022language} propose to place double tags on the encoder side, the capability of indicating the target language still remains insufficient \citep{wu-etal-2021-language}.
Therefore, this paper mainly focuses on the two most widely used strategies, {\em i.e,} T-Enc and S-Enc-T-Dec, to investigate the influence of the placement of LT.

\subsection{Off-Target issue in Zero-Shot Translation} \label{sec: back_off_target}
The \textit{off-target} issue describes the wrong target languages translated by the MNMT models on zero-shot translation \citep{zhang-etal-2020-improving}.
Prior studies \citep{gu-etal-2019-improved, zhang-etal-2020-improving, liu-etal-2021-improving-zero, wang-etal-2021-rethinking-zero, yang-etal-2021-improving-multilingual, mao2023exploring, zan2023unlikelihood} reveal that the spurious correlations between language pairs within supervised data aggravate this issue and make efforts to overcome it in terms of adjusting training strategy, modifying the residual connection and generating auxiliary data.
Besides, \citet{wang2022understanding} points out that data noises also make the \textit{off-target} issue, and \citet{wang2022understanding, jin2022informative, chen2023off} enhance the model's awareness to the vocabulary of target language during generation.
However, the cause of fundamental language tag strategies, which perform various on this issue, remains unclear.

\section{Probing Off-Target issue in MNMT}
In this section, we probe the \textit{off-target} issue in MNMT models with different LT strategies.
Firstly, we introduce the language rate/accuracy metric on this issue (\S\ref{sec: lang_acc}).
Secondly, we conduct experiments in terms of the distribution of this issue (\S\ref{sec: off_target}), the fine-grained language accuracy along decoding steps (\S\ref{sec: fine_graind_acc}) and the language indication in the encoder (\S\ref{sec: variation}).
Lastly, we expand our conclusion to propose \textbf{LCS} to mitigate this issue (\S\ref{sec: lcs}).

\subsection{Metrics of Off-Target issue} \label{sec: lang_acc}
To quantify this issue, we adopt the language rate as the metric to observe the error language distribution of zero-shot translation.
We adopt the langdetect\footnote{https://github.com/Mimino666/langdetect} toolkit to identify the language of generated sentences, following prior studies \citep{zhang-etal-2020-improving, wang-etal-2021-rethinking-zero, jin2022informative}.
And we calculate the language rate of language $\ell$ as follows:
\begin{equation}
    \mathit{Rate}(\ell) = \frac{\sum_{i=1}^{N}\mathbbm{1}_{lang(y^{(i)})=\ell}}{N} \times 100\%,
\end{equation}
where $\mathrm{y}^{(i)}$ denotes the $i$-th generated sentence, $N$ denotes the scale of the test set, and $lang(\cdot)$ denotes the language detect function.
The \textbf{accruacy} denotes the language rate of the desired target language, where the higher language accuracy denotes the slighter \textit{off-target} issue. 
Generally, language accuracy is regarded as an indicator of the performance of zero-shot translation, where the higher accuracy usually guides to the better performance.

To further observe the variation of language rate throughout decoding steps, we calculate the rate of continuous intervals with 5 words in it along generated sentences.
To accurately detect the language of short text, we choose the \textit{lingua-py}\footnote{https://github.com/pemistahl/lingua-py} toolkit, which has higher accuracy in detecting the short texts\footnote{More accurate as described as their GitHub repository.}.


\begin{table}[t!]
\centering
\resizebox{0.48\textwidth}{!}{
    \begin{tabular}{ c | c | c c c c}
    \toprule
    \textbf{Noise}? & \textbf{Strategy} & \cellcolor{ggreen}{\textbf{Acc} $\uparrow$} & \textbf{To-Src} $\downarrow$ & \textbf{To-En} $\downarrow$ & \textbf{To-Other} $\downarrow$ \\
    \midrule
    \multirow{4}*{\makecell[c]{Noise}} & T-Enc        & \cellcolor{ggreen}{44.38} & 38.71 & \;\;\textbf{3.62} & 13.29 \\
    ~ & S-Enc-T-Dec  & \cellcolor{ggreen}{14.09} & \;\;\textbf{0.44} & 72.46 & 13.01 \\
    ~ & ST-Enc       & \cellcolor{ggreen}{\;\;2.15} & \;\;3.71 & 81.96 & \textbf{12.18} \\
    ~ & ST-Enc-T-Dec & \cellcolor{ggreen}{\textbf{55.99}} & \;\;7.22 & 12.08 & 24.71 \\
    \midrule
    \multirow{4}*{\makecell[c]{De-\\Noise}} & T-Enc        & \cellcolor{ggreen}{\textbf{81.64}} & \;\;2.51 & \;\;\textbf{3.17} & \textbf{12.68} \\
    ~ & S-Enc-T-Dec  & \cellcolor{ggreen}{28.78} & \;\;\textbf{0.38} & 57.55 & 13.29 \\
    ~ & ST-Enc       & \cellcolor{ggreen}{36.06} & \;\;6.96 & 39.40 & 17.58 \\
    ~ & ST-Enc-T-Dec & \cellcolor{ggreen}{25.83} & \;\;0.69 & 52.40 & 21.08 \\
    \bottomrule
    \end{tabular}
}
\caption{
The average language rate (\%) of several LT strategies on the OPUS-100 dataset.
\colorbox{ggreen}{Acc}, To-Src, To-En, and To-Other denote the language rate of the expected target language, the source language, English, and other undesired languages in translation, respectively.
}
\label{tab: tag12}
\end{table}

\subsection{Error Distribution of Off-Target Issue} \label{sec: off_target}
In this section, we first conduct experiments on the OPUS-100 dataset, which contains 100 languages, to count the error distribution of the \textit{off-target} issue.
Prior study \citep{wang2022understanding} points out that the noise in data is an important factor of the \textit{To-Source} issue.
Thus, we also count the error distribution with the denoised data\footnote{Training details of both settings are shown in Appendix.\ref{appendix: training}.}, which filters the noise language pairs with target sentences in wrong languages, following \citet{wang2022understanding}.

As shown in Tab.\ref{tab: tag12}, the mainstream LT strategies suffer from the serious \textit{off-target} issue.
Most mainstream LT strategies endure the grave \textit{To-English} issue, which ranges from 39.40\% to 57.55\% even after data denoise, signifying that the language generation is disturbed by English.
Besides, the T-Enc and ST-Enc strategies tolerate the severer \textit{To-Source} issue than other strategies, indicating that the target tag on the encoder side may be mixed with the indication to the source language.
Error distributions on both data settings suggest that the placement of LT has a non-ignorable impact on the \textit{off-target} issue.
To understand the impact of LT better, we explore the language variation of language generation and the encoder's modeling tendency within different LT strategies.
Specifically, we employ the widely-used T-Enc and S-Enc-T-Dec strategies on the denoised OPUS-100 dataset to conduct analysis experiments.

\subsection{Fine-grained Language Accuracy along Decoding Steps} \label{sec: fine_graind_acc}
In this section, we explore the variation of language indication in generation, by calculating the fine-grained language rate throughout decoding steps. 

\begin{figure}[t!]
\begin{center}
    \scalebox{0.485}{
        \subfigure[] {
            \begin{minipage}[t]{\linewidth}
            \includegraphics[width=\textwidth]{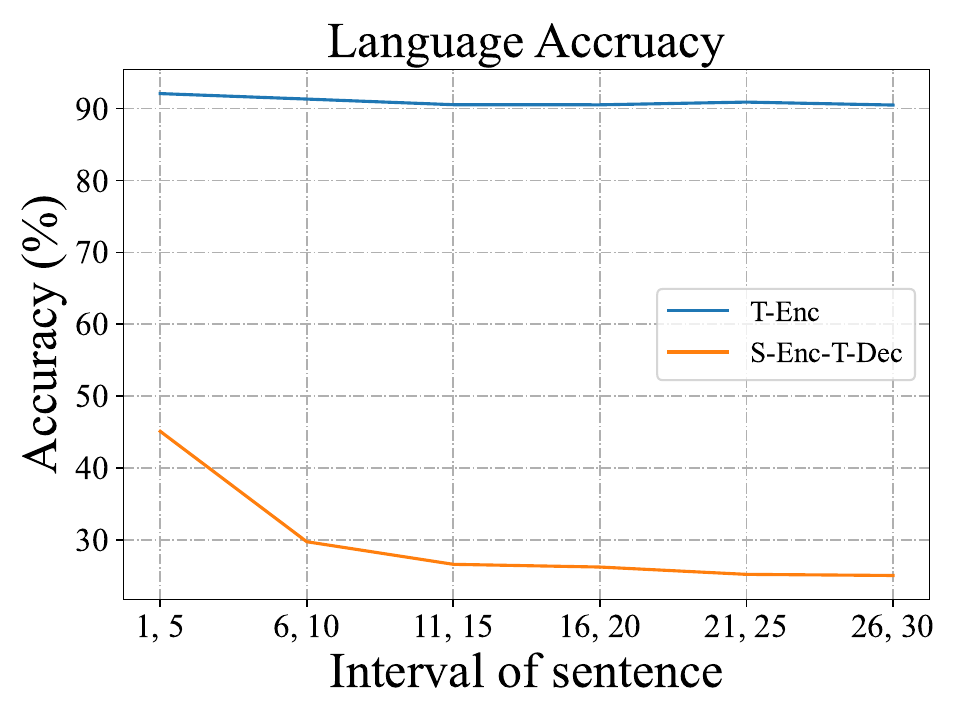}
            \end{minipage}
            \label{fig: opus_tgt_seg5}
        }
        \subfigure[] {
            \begin{minipage}[t]{\linewidth}
            \includegraphics[width=\textwidth]{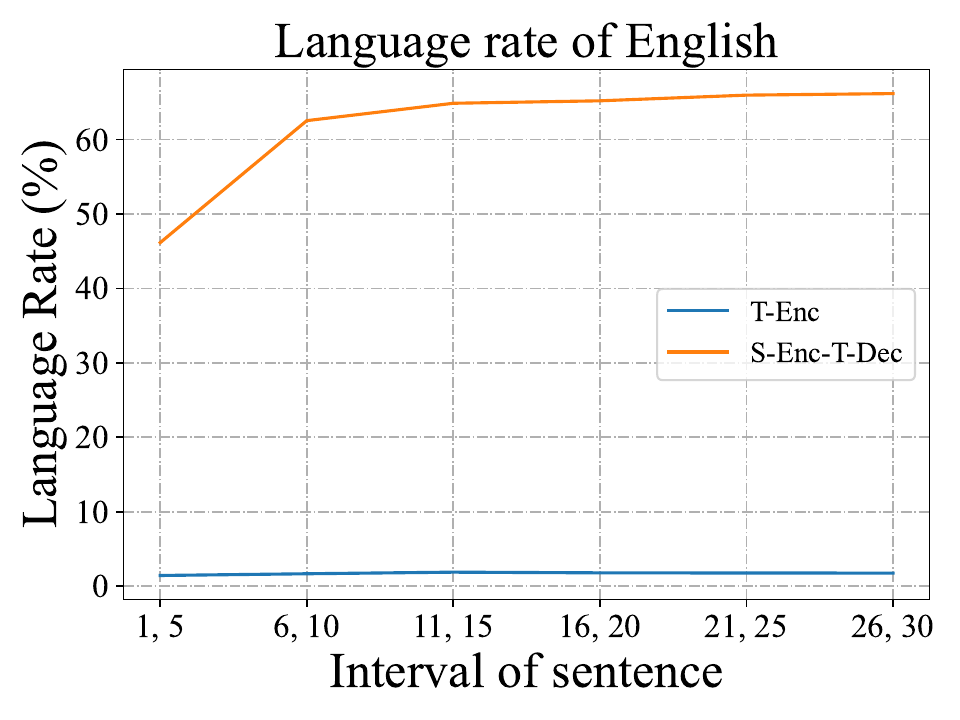}
            \end{minipage}
            \label{fig: opus_en}
        }
    }
    \caption{
      Fine-grained language rate of the desired target language (\emph{i.e.}, language accuracy) and undesired English throughout decoding steps on the zero-shot testset of denoised OPUS-100, with 5 words in each interval of the final translation results.
    } 
    \label{fig: decoding_steps}  
    \vspace{-4pt}
 \end{center} 
\end{figure}

As shown in Fig.\ref{fig: opus_tgt_seg5}, in terms of language accuracy, the S-Enc-T-Dec strategy exhibits a considerable decrease of approximately 20\%.
Besides, as shown in Fig.\ref{fig: opus_en}, the rate of mistranslating into English of S-Enc-T-Dec increases by around 20\%.
These variations signify that the indication from the S-Enc-T-Dec strategy is rapidly degraded and switched to English after generating a few tokens.
Hence, we conclude that the language indication of the S-Enc-T-Dec is decreasing after a few tokens, resulting in the bias to the most common language in the training set, \emph{i.e.}, English.
Conversely, the T-Enc strategy exhibits a more sufficient and stable language indication throughout the decoding steps.
Comparing both the above strategies, we conclude that the indication from the encoder could be more sufficient and stable and avoid being diminished during generation, while this setting suffers from the \textit{To-Source} issue.

\subsection{Language Indication in the Encoder} \label{sec: variation}
To invest the target language indication in the encoder and the \textit{To-Source} issue, we focus on two questions: 
1) \textit{How does the encoder model the target language tag and the input sentence of the source language, to obtain sufficient target language indication?}
2) \textit{Is there the encoder's language representation connected with the To-Source Issue?}

To answer the first question, we visualize the variation of language representation along the encoder layers, by calculating the similarity between the source and target languages in the zero-shot sentence pairs on the OPUS-100 dataset.
Following \citet{pan-etal-2021-contrastive}, we adopt the average-pooled encoder layer output as the sentence representation, and then calculate their cosine similarity.

\begin{figure}[t!]
\begin{center}
    \resizebox{0.375\textwidth}{!}{
        \includegraphics[width=1\textwidth]{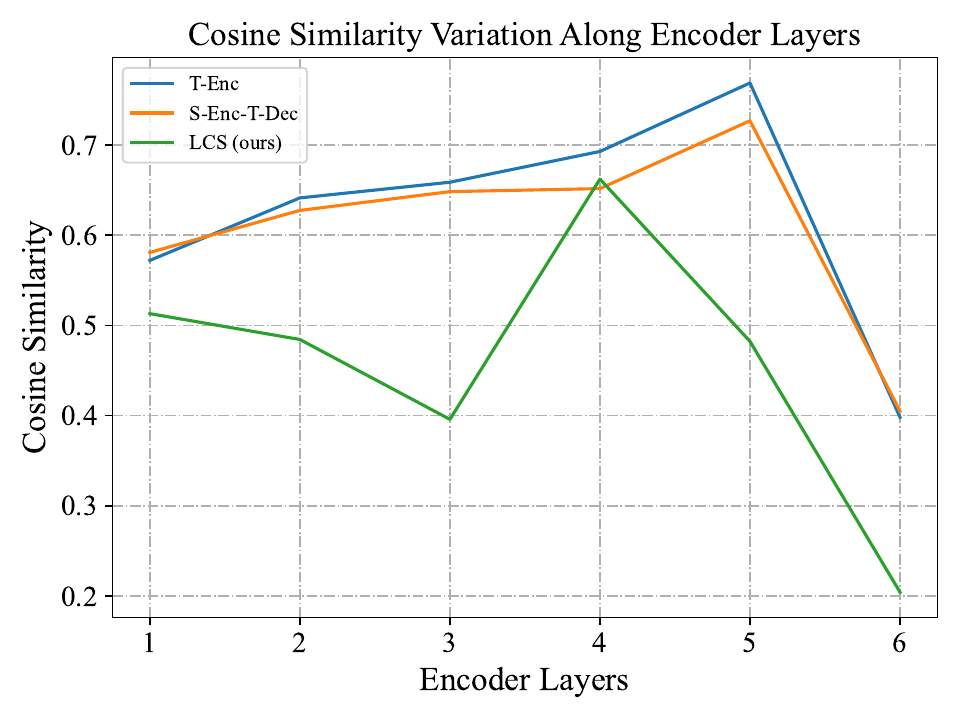}
    } 
    \caption{
        Curves of the similarity of the language pairs along encoder layers on the zero-shot testset of denoised OPUS-100. 
        The higher similarity denotes the representation is more similar and language-agnostic.
    } 
    \label{fig: variation}  
\end{center} 
\vspace{-4pt}
\end{figure}

As shown in Fig.\ref{fig: variation}, for the 6-layer encoder, the similarity scores of both LT strategies maintain the upward trend from the 1st layer to the 5th layer, and drop in the 6th layer, which is consistent with \citep{wu-etal-2021-language}.
The variation suggests that the encoder tends to generate language-agnostic representation from different languages first and then reduce the similarity across languages.
In fact, the representation of the top layer in the T-Enc strategy tends to be target-language-specific, whereas the one of the S-Enc-T-Dec strategy tends to be source-language-specific, which is verified in Appendix \ref{appendix: tsne}.
Hence, we could answer the first question that the encoder has a first-agnositc-second-specific tendency on language representation, the target LT mainly indicates the desired target language on the top encoder layer. 
Further, we conjecture that the target LT is mixed with the source language features in the first stage of the tendency, resulting in the \textit{To-Source} issue.

\begin{table}[t!]
\centering
\resizebox{0.48\textwidth}{!}{
    \begin{tabular}{ c | c | c  c  c  c }
    \toprule
    \textbf{Noise}? & \textbf{Strategy} & \cellcolor{ggreen}{\textbf{Acc}} $\uparrow$ & \textbf{To-Src} $\downarrow$ & \textbf{To-En} $\downarrow$ & \textbf{To-Other} $\downarrow$ \\
    \midrule
    \multirow{3}*{\makecell[c]{Noise}} &  
        T-Enc        & \cellcolor{ggreen}{44.38} & 38.71 & \;\;3.62 & 13.29 \\
    ~ & T-Enc-Mask      & \cellcolor{ggreen}{48.49} & 13.01 & 25.61 & 12.89 \\
    ~ & LCS (ours)   & \cellcolor{ggreen}{\textbf{85.35}} & \;\;\textbf{0.27} & \;\;\textbf{2.65} & \textbf{11.73} \\
    \midrule
    \multirow{3}*{\makecell[c]{De-\\Noise}} & T-Enc & \cellcolor{ggreen}{81.64} & \;\;2.51 & \;\;3.17 & 12.68 \\
    ~ & T-Enc-Mask & \cellcolor{ggreen}{78.62} & \;\;0.39 & \;\;9.62 & 11.37 \\
    ~ & LCS (ours)   & \cellcolor{ggreen}{\textbf{86.67}} & \;\;\textbf{0.19} & \;\;\textbf{2.08} & \textbf{11.06} \\
    \bottomrule
    \end{tabular}
}
\caption{The average language rate (\%) on the zero-shot testset of OPUS-100. 
T-Enc-Mask denotes the source LT is masked in each layer of 4 shallow encoder layers.
\colorbox{ggreen}{Acc}, To-Src, To-En, and To-Other denote the language rate of the expected target language, the source language, English, and other undesired languages in translation, respectively.
}
\vspace{-2pt}
\label{tab: tse}
\end{table}

To answer the second question and verify our conjecture, we apply a simple operation to the T-Enc strategy.
Specifically, we mask the target language tag in the 4 shallow encoder layers and restore it in the 5th layer\footnote{Our prior experiment shows that restoring the LT in the 5th layer performs better.
And we roughly implement it via masking and adding the initial embedding of the tag.}.
As shown in Tab.\ref{tab: tse}, this operation could significantly mitigate the \textit{To-Source} issue in the T-Enc strategy, reducing to 13.01\% ($\mathit{\Delta}$\textit{=-25.7\%}) and 0.39\% ($\mathit{\Delta}$\textit{=-2.12\%}) on this issue.
The remission on \textit{To-Source} issue verifies our conjecture and responds to the second question that the first-agnostic-second-specific tendency introduces bias into the target LT in the first stage.

In conclusion, we summarize our findings on the impact of the placement of LT on the \textit{off-target} issue as follows:
1) The language indication from the encoder side is more sufficient and stable, without being diminished during decoding; 
2) The target language is mainly indicated on the top encoder layers, and placing the target LT at the bottom layer of the encoder introduces the \textit{To-Source} issue.


\begin{figure}[t!]
\begin{center}
    \resizebox{0.475\textwidth}{!}{
        \includegraphics[width=1\textwidth]{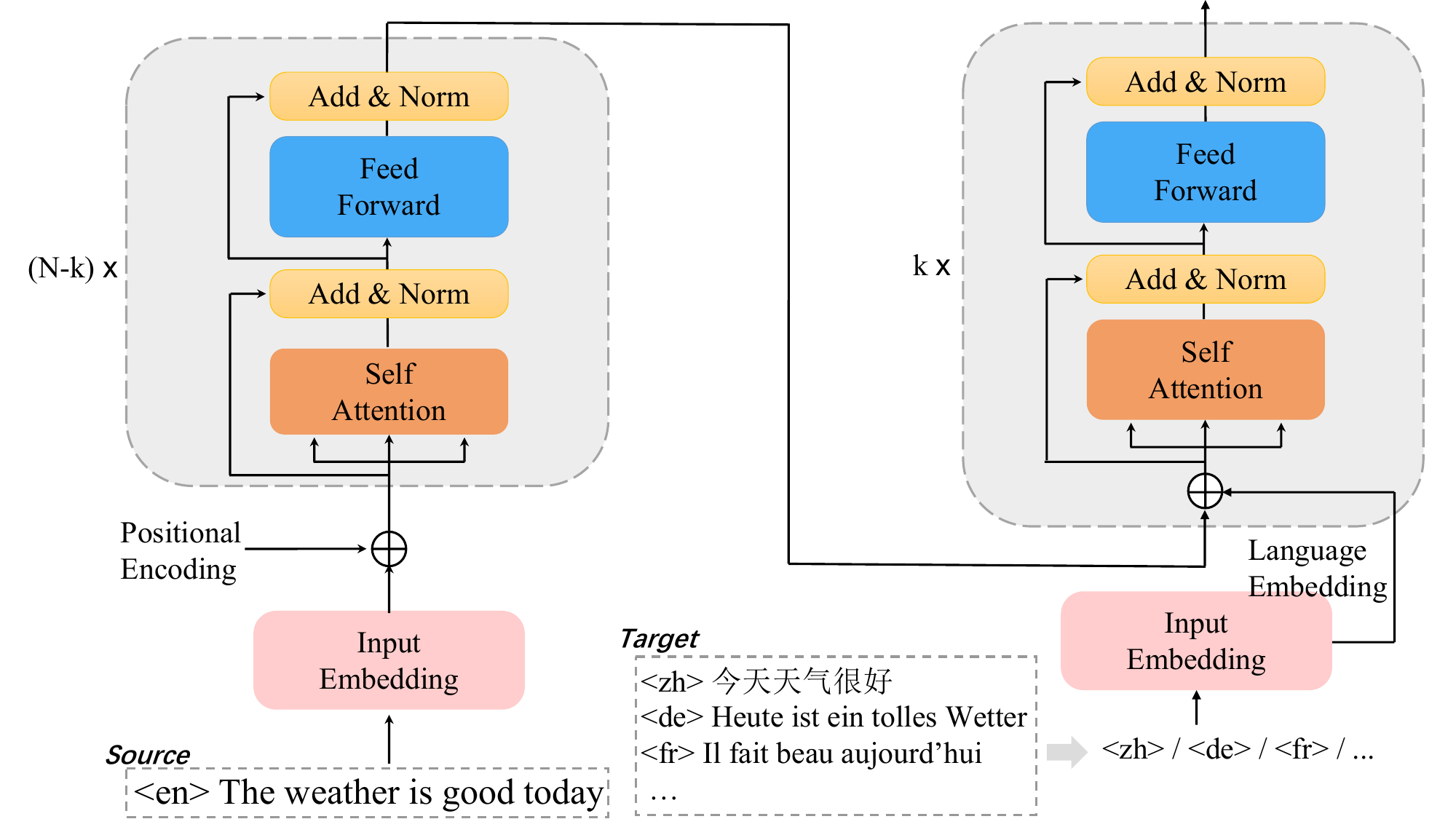}
    } 
    \caption{
        Illustration of the encoder of LCS.
        For the target, only the language tag could be seen by the encoder.
    } 
    \label{fig: lcs_method}  
\vspace{-6pt}
\end{center} 
\end{figure}

\subsection{Language Converter Strategy} \label{sec: lcs}
Based on the conclusions in \S\ref{sec: fine_graind_acc} and \S\ref{sec: variation}, we propose \textbf{L}anguage \textbf{C}onverter \textbf{S}trategy (\textbf{LCS}) to enhance the language indication to mitigate the \textit{off-target} issue, which further improves the quality of zero-shot translation.

\begin{table*}[t!]
\centering
\resizebox{\textwidth}{!} {
\begin{tabular}{c|ccc|ccc|ccc}
    \toprule
    \multirow{2}{*}{\textbf{Model}} & \multicolumn{3}{c|}{\textbf{MultiUN}} & \multicolumn{3}{c|}{\textbf{TED}} & \multicolumn{3}{c}{\textbf{OPUS-100}}\\
    ~ & \textbf{Supervised} & \textbf{Zero-Shot} & \cellcolor{ggreen}{\textbf{Accuracy}} & \textbf{Supervised} & \textbf{Zero-Shot} & \cellcolor{ggreen}{\textbf{Accuracy}} & \textbf{Supervised} & \textbf{Zero-Shot} & \cellcolor{ggreen}{\textbf{Accuracy}}\\
    \midrule
    T-Enc       & 50.73 & 32.96 & \cellcolor{ggreen}{92.58} & 25.20 & 10.41 & \cellcolor{ggreen}{94.50} & 24.72 / 22.81 & \;\;7.29 / 12.65 & \cellcolor{ggreen}{44.38 / 81.64} \\
    S-Enc-T-Dec & 50.71 & 23.63 & \cellcolor{ggreen}{71.85} & \textbf{25.27} & \;\;2.36 & \cellcolor{ggreen}{62.88} & 24.74 / 22.07 & \;\;3.80 / 4.71\;\; & \cellcolor{ggreen}{14.09 / 28.78} \\
    ST-Enc      & 50.66 & \;\;1.15 & \cellcolor{ggreen}{\;\;0.27} & 24.47 & \;\;1.37 & \cellcolor{ggreen}{13.66} & 24.82 / 22.82 & \;\;1.92 / 7.38\;\; & \cellcolor{ggreen}{\;\;2.15 / 36.06} \\
    ST-Enc-T-Dec & 50.66 & 19.95 & \cellcolor{ggreen}{64.15} & 24.51 & \;\;4.59 & \cellcolor{ggreen}{79.02} & \textbf{25.05} / 22.11 & \;\;4.23 / 3.60\;\; & \cellcolor{ggreen}{55.99 / 25.83} \\
    T-Enc-T-Dec & \textbf{50.83} & 32.23 & \cellcolor{ggreen}{91.10} & 24.53 & 12.07 & \cellcolor{ggreen}{95.73} & 24.75 / \textbf{22.89} & \;\;5.97 / 12.66 & \cellcolor{ggreen}{35.26 / 82.73} \\
    LCS (Ours)  & 50.75 & \;\textbf{36.03}* & \cellcolor{ggreen}{\;\textbf{95.28}*} & \textbf{25.27} & \;\;\textbf{13.71}* & \cellcolor{ggreen}{\;\textbf{96.21}*} & 24.80 / 22.09 & \textbf{15.22}* / \textbf{15.58}* & \cellcolor{ggreen}{\textbf{85.35}* / \textbf{86.67}*} \\
    \bottomrule
\end{tabular}
}
\caption{
    Experiments in several LT strategies on the MultiUN, TED, and OPUS-100 datasets. 
    Supervised and Zero-Shot denote the average BLEU scores of the supervised and zero-shot translation directions. 
    \colorbox{ggreen}{Accuracy} denotes the averaged language accuracy (\%) of zero-shot translation. 
    "A / B" separates the scores of noise data and denoise data in OPUS-100, where `A' and `B' represent the result of the noise and denoised version, respectively.
    Results with */** are statistically better than `T-Enc' in all translation directions with $p < 0.01$.
    \textbf{Bold} denotes the best performance.
}
\label{tab: main_results}
\end{table*}

According to our conclusions, as shown in Fig~\ref{fig: lcs_method}, we divide the encoder layers into the shallow layers and the deep language converter layers by the $k$-th\footnote{$k$ is a hyperparameter, describing the number of the language converter layers. 
We set it to 2 in the 6-layer encoder and explore the selection of $k$ in \S\ref{sec: hyperk}.} deepest layer.
In the shallow layers, we place the source language tag in front of the sentences to avoid the \textit{To-Source} issue.
In the deep language converter layers, we introduce the target language embedding as auxiliary signals to prompt the desired target language.
%
As pointed out by prior studies~\citep{bjerva-etal-2019-language, oncevay-etal-2020-bridging, jin2022informative}, the target language embeddings contain target language-specific features, which could provide sufficient indication of the target language into the top encoder layers.

Then the final calculation of the self-attention block \citep{vaswani2017attention} in language converter layers is as follows:
\begin{equation}
    \Tilde{h}_i = h_i + e^{t},
\end{equation}
\begin{equation}
    s = LayerNorm(\Tilde{h} + SelfAttn(\Tilde{h})),
\end{equation}
where $h_i$ denotes the $i$-th state of input tokens to each converter layer, $s$ denotes the output states of self-attention block in each converter layer, $SelfAttn$ denotes the calculation of self-attention and $LayerNorm$ represents the LayerNorm function. 
Since the language tags have already been included in the vocabulary, LCS introduces no extra parameters.
Besides, we maintain the cross-entropy loss to optimize the MNMT model.

Moreover, we place the target language tag in front of the decoder input to indicate the target language better.
We verify the effectiveness of this placement in Appendix \ref{sec: placement}.

\begin{table}[t!]
\centering
\resizebox{0.475\textwidth}{!}{
\begin{tabular}{c | c  l | r  }
    \toprule
    \multirow{2}*{\textbf{Dataset}} & \multicolumn{2}{c|}{\textbf{Langs \& Dirs}} & \multirow{2}*{\textbf{Train / Valid / Test}} \\
    ~ & \textbf{Supervised} & \textbf{Zero-Shot} & ~ \\
    \midrule
    MultiUN & 4\;\; \& \,\;6 & \;3\; \& \;\,6\;\;\, & 2M\;\ \quad / 4K / 4K  \\
    TED & 20\; \& \,38 & 19 \& 342 & 14K-22K / 5K / 5K \\
    OPUS-100 & 100 \,\& 198 & \;6\; \& \;30 & 10K-1M\; / 2K / 2K \\
    \bottomrule
\end{tabular}
}
\caption{
    Data statistics. 
    Langs\&Dirs represents the number of languages and translation directions involved in the supervised and zero-shot translation.
    Train / Valid / Test represents the number of samples in each translation direction in the training / validation / test set.
}
\label{tab: datasets}
\vspace{-4pt}
\end{table}

\section{Experiments}
\subsection{Experimental Setup}
We conduct experiments on three popular datasets, MultiUN, TED, and OPUS-100 \citep{gu-etal-2019-improved, qi-etal-2018-pre, zhang-etal-2020-improving}.
%
The statistics of the datasets of each translation direction are presented in Tab.\ref{tab: datasets}. 
(Please refer to Appendix~\ref{appendix: training} for more details.)
For direct comparison, we report scores of noise and denoised data versions of OPUS-100.
For all these datasets, English is the central language in the training sets, serving as either the source or the target language in each sentence pair. 
Following \citet{johnson-etal-2017-googles}, we consider translation directions involving English as supervised translation and directions between non-English languages as zero-shot translation.

We employ open-source toolkit fairseq\citep{ott2019fairseq} to implement the Transformer models, with mixed precision \citep{ott-etal-2018-scaling}.
During inference, we set beam size to 5 and length penalty to 1.0, following existing studies \citep{wang-etal-2021-rethinking-zero, jin2022informative}.
We apply SacreBLEU to calculate BLEU and report the averaged scores for all models.
More details about the dataset, training, and evaluation can be found in Appendix \ref{appendix: training}.

\subsection{Main Results}
We respectively establish MNMT models in the several LT strategies in \S\ref{sec: lt_strategy} and our LCS on the three datasets and list the results in Tab.\ref{tab: main_results}. 
Besides, we conduct experiments in the T-Enc-T-Dec, which seems to combine the indication from both encoder and decoder sides.
Moreover, we also report the results of the denoised OPUS-100 dataset for a comprehensive comparison, where MultiUN and TED are low-noise datasets.

As shown in Tab.\ref{tab: main_results}, our proposed LCS yields the highest language accuracy and best performance of zero-shot translation on all datasets among all LT strategies.
Specifically, compared to the widely-used T-Enc strategy, LCS effectively mitigates the \textit{off-target} issue by improving language accuracy up to 95.28\% (\textit{+2.7\%}), 96.21\% (\textit{+1.71\%}), 85.35\% (\textit{+40.97\%}), and 86.67\% (\textit{+5.03\%}) on zero-shot translation of MultiUN, TED, OPUS-100 (noise and denoised) datasets, respectively.
Furthermore, LCS outperforms the T-Enc strategy by 3.07, 3.30, 7.93, and 2.93 BLEU scores improvements on zero-shot translation of these datasets, respectively. 
While bringing conspicuous improvement to zero-shot translation, LCS maintains the performance of supervised translation.
%
Unfortunately, since the language detection toolkit is lowly accurate for medium- and low-resource languages, the performance on related translation pairs is reduced in the denoised OPUS-100.
In this case, LCS performs well and robust on zero-shot translation on the noise version of the OPUS-100 dataset, compared to the denoise version.

We also probe the performance of our method in prior aspects, \emph{i.e.}, the distribution of the \textit{off-target} issue, and the language variation in the encoder.
As shown in Tab.\ref{tab: tse}, LCS could yield the highest language accuracy, and the lowest rates on the \textit{To-Source} and \textit{To-English} issue.
These scores suggest that LCS could provide the most accurate target language indication for zero-shot translation, and avoid being mixed with the indication of source language.
Since LCS performs better than T-Enc on the \textit{To-English} issue, we consider that LCS also provides sufficient and stable indication during generation.
Besides, in terms of language representation variation, our proposed LCS yields the lowest score in the 6-th layer than the T-Enc strategy, meaning that LCS could generate more target-language-specific representation to indicate the target language better.

\section{Analysis}
In this section, we explore the details of LCS to understand it better.
We invest the generalizability of LCS to other approaches (\S\ref{sec: app}) and the effect of hyperparameter $k$ (\S\ref{sec: hyperk}), and the application of LCS to deeper encoders (\S\ref{sec: deep_enc}).

\subsection{Application to Stronger Approaches} \label{sec: app}
We evaluate the generalizability of LCS on OPUS-100 by applying it to the following stronger approaches:

\begin{table}[t!]
\centering
\resizebox{0.475\textwidth}{!}{
    \begin{tabular}{ c | c  c  c }
    \toprule
    \textbf{Model} & \textbf{Supervised} & \textbf{Zero-Shot} & \cellcolor{ggreen}{\textbf{Accuracy}} \\ 
    \midrule
    S-Enc-T-Dec & 24.74 / 22.07 & \;\;3.80 / \;\;4.71 & \cellcolor{ggreen}{14.09 / 28.78} \\
    FT \& LCS & \textbf{25.29} / \textbf{22.34} & \textbf{15.45} / \textbf{15.90} & \cellcolor{ggreen}{\textbf{85.26} / \textbf{86.32}} \\
    \midrule
    DisPI & 24.62 / \textbf{22.09} & \;\;4.80 / \;\;4.94 & \cellcolor{ggreen}{18.91 / 30.36} \\
    DisPI \& LCS & \textbf{24.78} / 21.87 & \textbf{15.71} / \textbf{15.80} & \cellcolor{ggreen}{\textbf{85.34} / \textbf{87.35}} \\
    \midrule
    DN\dag & - & - & - \\
    DN & 19.04 / \textbf{22.83} & \;\;3.02 / 12.62 & \cellcolor{ggreen}{21.75 / 81.66} \\
    DN \& LCS & \textbf{24.32} / 22.10  & \textbf{14.00} / \textbf{15.55} & \cellcolor{ggreen}{\textbf{81.81} / \textbf{86.60}} \\
    \midrule
    LEE\dag & \textbf{24.98} / - & 10.08 / - &\cellcolor{ggreen}{79.90 / -}\\
    LEE & 24.13 / 21.56 & 11.88 / 12.30 & \cellcolor{ggreen}{73.69 / 80.00} \\
    LEE \& LCS & 24.88 / \textbf{21.91} & \textbf{15.20} / \textbf{15.63} & \cellcolor{ggreen}{\textbf{85.91} / \textbf{86.97}} \\
    \midrule
    CTS & \textbf{24.21} / \textbf{21.72} & 12.77 / 10.34 & \cellcolor{ggreen}{80.09 / 62.35} \\
    CTS \& LCS & 24.19 / 21.52 & \textbf{15.13} / \textbf{15.43} & \cellcolor{ggreen}{\textbf{86.85} / \textbf{87.86}} \\
    \midrule
    \midrule
    mBART \& FT\;\;  & 29.03 / 29.53 & 3.47 / 4.94 & \cellcolor{ggreen}{5.10 / 13.05} \\
    mBART \& LCS & \textbf{30.84} / \textbf{31.33} & \textbf{21.15} / \textbf{21.24} & \cellcolor{ggreen}{\textbf{86.14} / \textbf{87.51}} \\
    \bottomrule
    \end{tabular}
}
    \caption{
        Experiments about the application of LCS to other approaches.
        `\dag' represents that the results are cited from the corresponding papers, and the rest models are reproduced by us.
        "/" separates the scores of noise data and denoise data in OPUS-100.
    }
\label{tab: comparison}
\end{table}
\begin{itemize}[leftmargin=*,topsep=0pt]
\setlength{\itemsep}{0pt}
\setlength{\parsep}{0pt}
\setlength{\parskip}{0pt}
    \item \textbf{Fine-Tune (FT)}.
    We first train the model in the S-Enc-T-Dec strategy with 100K steps, and fine-tune the model in our strategy, with the same total training steps as other models.
    \item \textbf{Denosing Encoder (DN)} \citep{wang-etal-2021-rethinking-zero}.
    DN introduces the denoising auto-encoder training objective to bridge the connection between zero-shot language pairs.
    \item \textbf{Disentangling Positional Information (DisPI)} \citep{liu-etal-2021-improving-zero}.
    DisPI removes the residual connection of the encoder middle layer to yield the language-agnostic representation.
    \item \textbf{Contrastive Learning (CTS)} \citep{pan-etal-2021-contrastive}.
    CTS introduces the contrastive learning training objective to close the representation gap of similar sentences.
    We apply this objective to our shallow encoder layers since the function is similar to the first encoder stage of LCS.
    \item \textbf{Language Embedding Embodiment (LEE)} \citep{jin2022informative}.
    LEE adopts the target language embedding added to each state at the decoder side to indicate the desired target language without any LT strategies.
    \item \textbf{mBART} \citep{liu-etal-2020-multilingual-denoising}.
    mBART is a widely-used pretrained multilingual sequence-to-sequence model.
    We finetune mBART in vanilla S-Enc-T-Dec and LCS strategies, with the same training setting.
    Training setting of mBART is a bit different from others, details can be referred in Appendix~\ref{appendix: training}.
\end{itemize}

\noindent  \textbf{Results}. 
As shown in Tab.\ref{tab: comparison}, LCS is effective in enhancing the performance of existing approaches.
Specifically, with regard to zero-shot translation and language accuracy, most approaches achieve significant improvements with the application of LCS. 
%
%
On the OPUS-100 dataset, all approaches yield significant improvements with the application of LCS, with 11.65, 10.91, 10.98, 3.32, and 2.36 average BLEU scores improvements on the noise data version, respectively.
Compared to the noise version of the OPUS-100 dataset, LCS also could yield similar improvements on the denoise version, even achieving the 15.90 BLEU score by the FT method and the 87.86\% accuracy by the CTS\&LCS method.
Further, on the pretrained multilingual model, LCS exhibits advanced performance to enhance the performance of mBART on both supervised and zero-shot translation.
While in terms of supervised translation, the application of LCS generally maintains or slightly improves the performance of most approaches.
In conclusion, these results demonstrate that LCS is well-compatible with other approaches and can achieve further improvement when combined with them, without introducing extra parameters.

\begin{figure}[t!]
\begin{center}
    \resizebox{0.475\textwidth}{!}{
        \includegraphics[width=1\textwidth]{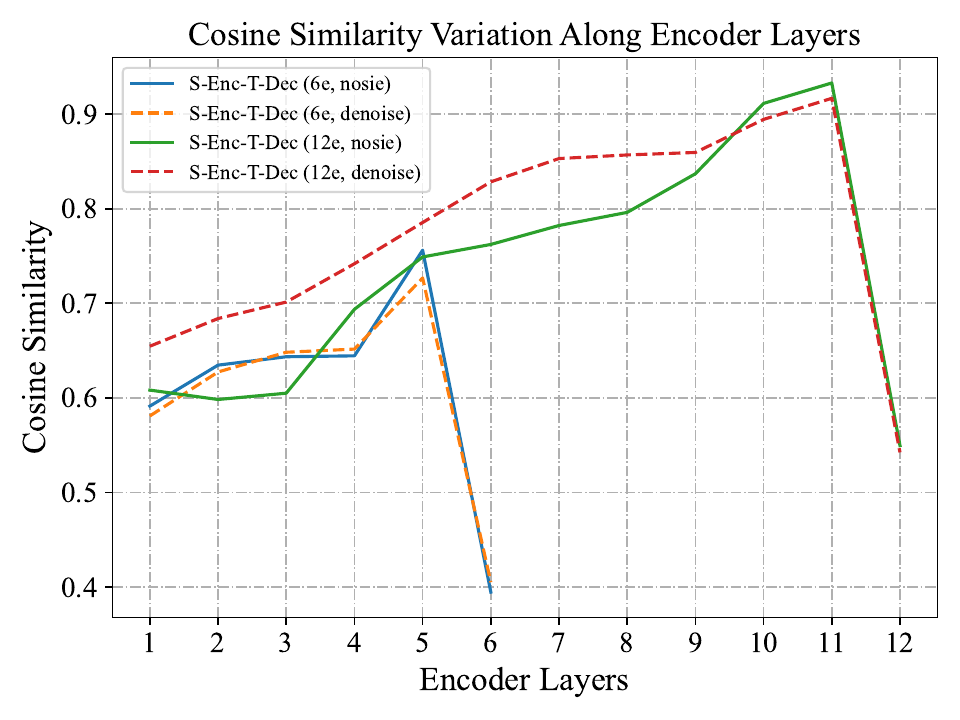}
    } 
    \caption{
        Curves of the similarity of the language pairs along encoder layers in S-Enc-T-Dec on the zero-shot testset of noise and denoised OPUS-100. 
    } 
    \label{fig: variation_12e}  
\vspace{-6pt}
\end{center} 
\end{figure}
\begin{table}[t!]
\centering
\resizebox{0.475\textwidth}{!}{
    \begin{tabular}{c | c | c | c | c }
    \toprule
    \multirow{3}*{\textbf{$\mathbf{k}$}} & \multicolumn{2}{c|}{\textbf{Supervised}} & \multicolumn{2}{c}{\textbf{Zero-Shot}} \\
    \cmidrule{2-5}
    ~ & \textbf{6-layer} & \textbf{12-layer} & \textbf{6-layer} & \textbf{12-layer} \\
    \midrule
    2 & \textbf{24.80} / \textbf{22.09} & \textbf{26.62} / \textbf{23.40}  & 15.22 / \textbf{15.58} & 16.86 / \textbf{17.39} \\
    3 & 24.64 / 21.76 & 26.60 / 22.85 & \textbf{15.32} / 15.51 & \textbf{17.01} / 16.86 \\
    4 & 24.13 / 21.22 & \;\;2.62 / 9.03\;\; & 15.06 / 15.06 & \;\;2.33 / 4.29\;\; \\
    5 & 23.20 / 20.81 & 23.75 / 19.93 & 14.03 / 14.75 & 15.40 / 14.62 \\
    6 & 23.62 / 20.99 & 22.70 / 20.15 & 12.79 / 14.22 & 14.45 / 14.66 \\
    \bottomrule
    \end{tabular}
}
    \caption{
        BLEU scores of LCS with different $k$ on the noise and denoise OPUS-100.
        $k$ denotes the hyperparameter $k$ and 6/12-layer denotes the encoder's depth.
    }
\label{tab: lt_layers}
\end{table}


\subsection{Selection of Hyperparameter $k$} \label{sec: hyperk}
The selection of hyperparameter $k$ is an important factor of LCS.
Indeed, the selection of $k$ mainly relies on the variation of language similarity among encoder layers, which is described in Section~\ref{sec: variation}.
We expand the variation into the deeper encoder with 12 layers, and display the variation in Fig.\ref{fig: variation_12e}.
We list the performance of different values of $k$ in the 6-layer and 12-layer encoder in Tab.\ref{tab: lt_layers}.

As the results show, the optimal selection of $k$ is around the inflection point of the variation of language similarity.
We could observe from Fig.\ref{fig: variation_12e} and Tab.\ref{tab: lt_layers}, when the selected value $k$ is around the inflection point, LCS could yield better translation quality.
For most settings of $k$, the performance of LCS on zero-shot translation is much better than other methods (as shown in Tab.\ref{tab: main_results} and Tab.\ref{tab: comparison}).
We further probe the selection of $k$ in other deeper encoders (24-layer and 48-layer), and conclude that the better range selection of $k$ is the nearby integers of 15\% of the encoder depth.

\subsection{Effect of LCS on deeper encoders} \label{sec: deep_enc}
\begin{figure}[t!]
\begin{center}
    \scalebox{0.485}{
        \subfigure[] {
            \begin{minipage}[t]{\linewidth}
            \includegraphics[width=\textwidth]{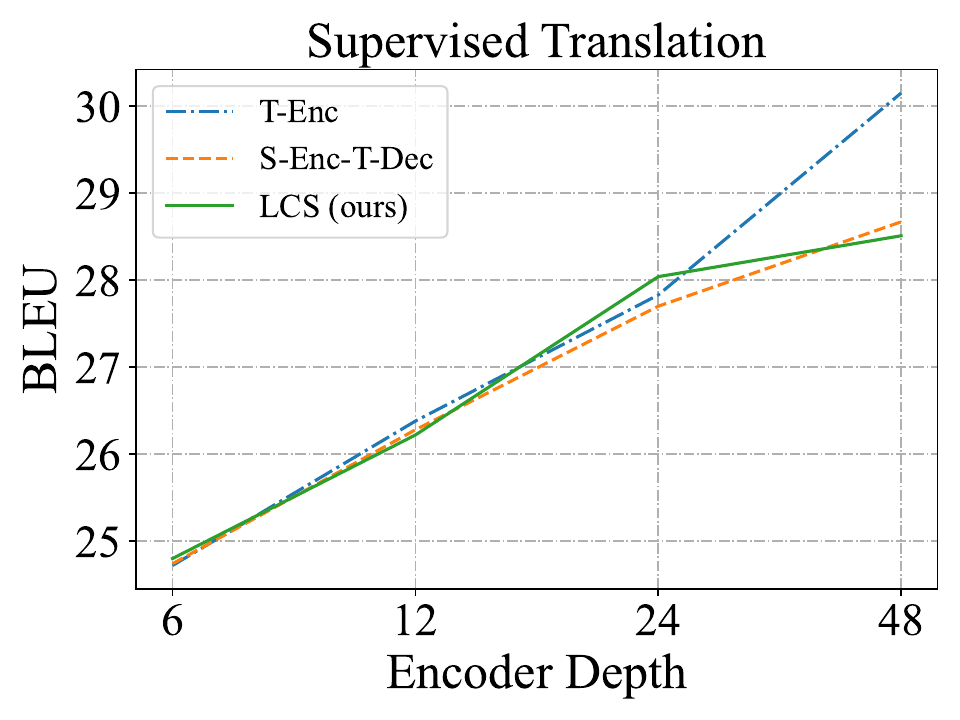}
            \end{minipage}
            \label{fig: supervised_trans}
        }
        \subfigure[] {
            \begin{minipage}[t]{\linewidth}
            \includegraphics[width=\textwidth]{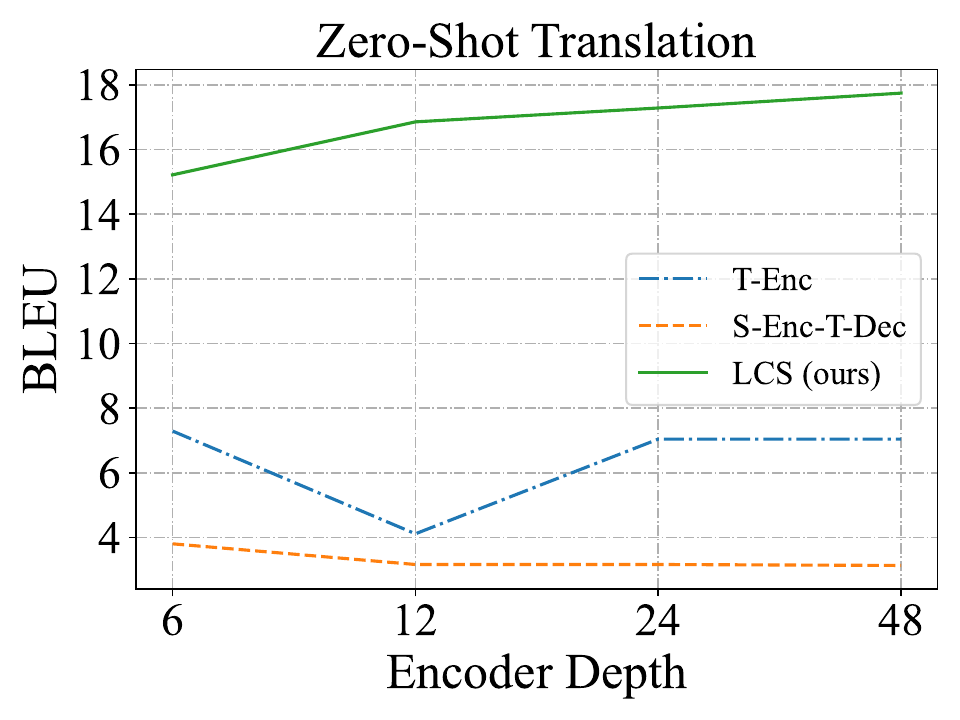}
            \end{minipage}
            \label{fig: zeroshot_trans}
        }
    }
    \caption{
    Performance of deeper encoder on the supervised and zero-shot testset of the noise OPUS-100.
    } 
    \label{fig: deep_trans}  
 \end{center} 
\end{figure}

We conduct experiments to compare the effectiveness of LCS with the T-Enc and S-Enc-T-Dec, where we set $k$ to 2 for 6- and 12-layer encoders, 5 for the 24-layer encoder, and 6 for the 48-layer encoder\footnote{In spired by \citet{wang2022deepnet}, we train 24- and 48-layer encoder with the assistance of DeepNorm.}.
As shown in Fig.\ref{fig: deep_trans}, although T-Enc and S-Enc-T-Dec could yield better translation on supervised translation with deeper encoders, both cannot perform much better than the 6-layer encoder on zero-shot translation.
Compared to them, with deeper encoders, LCS exhibits an upward trend on zero-shot translation, where the 48-layer encoder improves around 2.50 BLEU score compared to the 6-layer encoder.
Besides, in deep encoders, LCS maintains a similar supervised performance with the S-Enc-T-Dec strategy.

\section{Conclusion}
In this paper, we have identified that the \textit{off-target} issue is sensitive to the placement of LT and provide the analysis for the details of the issue with the prevalent LT strategies.
We reveal that placing LT on the encoder side provides a more stable and sufficient language indication than the decoder side, and introducing the target information into the top layers of the encoder will mitigate the confusion between the source and the target language.
Based on our findings, we propose LCS to mitigate the \textit{off-target} issue, and further improve the performance of zero-shot translation.
Extensive experiments and analysis suggest that LCS boosts the performance of zero-shot translation and significantly mitigates the \textit{off-target} issue without introducing extra parameters. Besides, LCS could perform well in the noised data set.

\section*{Limitations}
Compared to the boosted performance on zero-shot translation, LCS yields limited improvements on the supervised translation, while it is designed to enhance the language indication for zero-shot translation.
We intend to explore ways to improve the performance of zero-shot translation further and achieve greater improvement in supervised translation.

\section*{Acknowledgements}
The research work described in this paper has been supported by the National Nature Science Foundation of China (No. 61976016, 62376019, 61976015), and the authors would like to thank the anonymous reviewers for their valuable comments and suggestions to improve this paper.

\bibliography{anthology,custom}
\bibliographystyle{acl_natbib}

\appendix
\section{Dataset \& Training details} \label{appendix: training}
We show the detail of the MultiUN\footnote{https://conferences.unite.un.org/UNCORPUS}, TED\footnote{https://github.com/neulab/word-embeddings-for-nmt}, and OPUS-100\footnote{https://opus.nlpl.eu/opus-100.php} as the following sections.
We average the last five checkpoints to form the final tested checkpoint for all models in our experiments.

\subsection{MultiUN}
We select four languages distributed in various language families, Arabic (Ar), English (En), Russian (Ru), and Chinese(Zh), following \citet{wang-etal-2021-rethinking-zero}.
In final, the training set consists of 6M sentence pairs.
We select Transformer-base architecture based on post-norm to conduct experiments, and set 6 encoder/decoder layers with 8 attention heads, embedding size of 512, inner size of 2048, the dropout rate of 0.1, the maximum learning rate of 0.0007 and label smoothing rate of 0.1.
We share the vocabulary for all languages and segment words into subwords using byte pair encoding (BPE) \citep{sennrich-etal-2016-neural} with 40k merge operations, following \citet{wang-etal-2021-rethinking-zero}. 
In training, we set the maximum batch size per GPU to 4096 tokens and trained on 8 GPUs with 300K steps.

\subsection{TED}
It includes 60 languages in total \citep{qi-etal-2018-pre} and we choose the top 20 languages following \citet{qu2022adapting}.
In final, the training set consists of 3.5M sentence pairs.
We choose Transformer-base architecture based on post-norm to conduct experiments, and set 6 encoder/decoder layers with 8 attention heads, embedding size of 512, inner size of 2048, dropout rate of 0.1, maximum learning rate of 0.0005 and label smoothing rate of 0.1.
In this dataset, we use SentencePiece to segment words into subwords with 64k merge operations, following \citet{qu2022adapting}.
In training, we set the maximum batch size per GPU to 6400 tokens and trained on 8 GPUs with 100K steps.

\subsection{OPUS-100}
It includes 100 languages in total and consists of 55M training sentence pairs with up to 1M samples per language pair.
We choose Transformer-base architecture based on post-norm to conduct experiments, and set 6 encoder/decoder layers with 8 attention heads, embedding size of 512, inner size of 2048, dropout rate of 0.1, maximum learning rate of 0.0005 and label smoothing rate of 0.1.
In this dataset, we use SentencePiece \citep{kudo-richardson-2018-sentencepiece} to segment words into subwords with 64k merge operations, following \citet{jin2022informative}.
In training, we set the maximum batch size per GPU to 6400 and trained on 8 GPUs with 400K steps.
We train models in both noise and denoise data versions with the same parameter settings and similar epochs.

\subsection{mBART}
Since the number of languages in mBART is less than OPUS-100, we mainly select the six languages in the zero-shot test set of OPUS-100 (\emph{i.e.}, Arabic, German, French, Russian, and Chinese), and add English to conduct experiments.
The training set is still English-centric.
We select the mBART-Large\footnote{\url{https://dl.fbaipublicfiles.com/fairseq/models/mbart/mbart.cc25.v2.tar.gz}} with 25 languages.
During training, we set dropout to 0.3, and learning rate to 0.00003 with 2500 steps to warmup.
We set the maximum batch size per GPU to 1024 tokens and trained on 8 GPUs with 100K steps.
For fine-tuning mBART in LCS strategies, we set $k$ to 2.

\subsection{Deep Encoder Training}
Inspired by the successful application of DeepNorm \citep{wang2022deepnet} and BranchNorm \citep{liu2023branchnorm}, we utilize DeepNorm to stabilize the training of deep models (24-layer and 48-layer), which applies such a constraint to the early stage of model training.
And the $\alpha$ and $\beta$ are following DeepNorm, as the following formulas:
\begin{equation}
\begin{split}
\begin{aligned}
    \alpha_{encoder} &= 0.81 (N^4M)^{\frac{1}{16}}, \\
    \beta_{encoder} &= 0.87 (N^4M)^{-\frac{1}{16}}, \\
    \alpha_{encoder} &= (3M)^{\frac{1}{4}}, \\
    \beta_{decoder} &= (12M)^{-\frac{1}{4}},
\end{aligned}
\end{split}
\end{equation}
where $N$ and $M$ denote the depth of the encoder and decoder for a standard Transformer.


\subsection{Evaluation}
In MultiUN, we calculate the case-insensitive sacreBLEU scores, following \citet{wang-etal-2021-rethinking-zero}, and calculate case-sensitive sacreBLEU scores for TED and OPUS-100 datasets, following \citet{jin2022informative}.
We respectively average the scores of overall supervised or zero-shot translation directions to report in tables.
Specifically, we apply different tokenizer for all Chinese testset\footnote{SacreBLEU signatures: BLEU+case.mixed+numrefs.1\\+smooth.exp+tok.zh+version.2.0.0.}, compared to the rest languages\footnote{SacreBLEU signatures: BLEU+case.mixed+numrefs.1\\+smooth.exp+tok.13a+version.2.0.0.}.

\begin{table}[t!]
\centering
\resizebox{0.435\textwidth}{!}{
    \begin{tabular}{c | c | c  c  c }
    \toprule
    \textbf{Dataset} & \makecell[c]{\textbf{Strategy}} & \textbf{Supervised} & \textbf{Zero-Shot} & \cellcolor{ggreen}{\textbf{Accuracy}} \\ 
    \midrule
    \multirow{6}*{\makecell[c]{MultiUN\\(Small)}} & sS-cS & \textbf{50.75} & \textbf{36.03} & \cellcolor{ggreen}{\textbf{95.28}} \\
    ~ & sS-cT & \textbf{50.75} & 35.51 & \cellcolor{ggreen}{94.69} \\
    ~ & sS-c\_ & 50.73 & 35.93 & \cellcolor{ggreen}{95.12} \\
    ~ & s\_-c\_ & 50.63 & 35.76 & \cellcolor{ggreen}{95.04} \\
    ~ & sT-cT & 34.68 & 11.88 & \cellcolor{ggreen}{18.53} \\
    \cmidrule{2-5}
    ~ & Remove & 50.83 & 35.86 & \cellcolor{ggreen}{95.17} \\
    \midrule
    \multirow{6}*{\makecell[c]{OPUS-100\\(Large)}} & sS-cS & 24.80 / 22.09 & \textbf{15.22} / \textbf{15.58} & \cellcolor{ggreen}{85.35 / 86.67} \\
    ~ & sS-cT & 24.82 / \textbf{22.51} & 15.18 / 15.56 & \cellcolor{ggreen}{85.28 / 86.95} \\
    ~ & sS-c\_ & \textbf{24.83} / 22.15 & 15.14 / 15.56 & \cellcolor{ggreen}{85.08 / 86.79} \\
    ~ & s\_-c\_ & 23.84 / 21.63 & 14.43 / 15.10 & \cellcolor{ggreen}{\textbf{85.95} / \textbf{87.02}} \\
    ~ & sT-cT & 24.77 / 22.08 & \;\;5.37 / 12.36 & \cellcolor{ggreen}{29.56 / 79.11} \\
    \cmidrule{2-5}
    ~ & Remove & 24.73 / 21.96 & 12.22 / 13.78 & \cellcolor{ggreen}{52.44 / 74.95} \\
    \bottomrule
    \end{tabular}
}
    \caption{
        Experiments of the LCS strategy variants. `s/c' denotes the shallow stage or the language converter stage in the encoder of LCS, `S/T' denotes the source or target LT in front of sentences, and `\_' denotes no LT in this stage.
    }
\label{tab: placement}
\end{table}

\section{Impact of the Language Tag Placement} \label{sec: placement}
In this section, we investigate the impact of the LT placement for LCS.
We first conduct experiments to explore the influence of the source and target LT in two stages of LCS.
As shown in Tab.\ref{tab: placement}, when placing the source language tag on the shallow stage, the choice of language tag on the language converter stage will yield a relatively small impact on both datasets.
However, placing the target LT or no placing performs unstable on the shallow stage, with bad BLEU scores.
We also explore the impact of removing the target LT on the decoder side on LCS, as listed as `Remove' in Tab.\ref{tab: placement}. 
Compared to the vanilla LCS strategy (`sS-cS' in Tab.\ref{tab: placement}), `Remove' degrades 0.17 and 3.00 BLEU scores on zero-shot translation of two datasets, suggesting that placing the target LT on the decoder side is beneficial for LCS to perform better on zero-shot translation.
Therefore, we conclude that placing the source LT on the encoder side and the target LT on the decoder side is optimal for LCS.

\begin{table}[t!]
\centering
\resizebox{0.465\textwidth}{!}{
    \begin{tabular}{c | c | c | c | c }
    \toprule
    \multirow{3}*{\textbf{$\mathbf{k}$}} & \multicolumn{2}{c|}{\textbf{Supervised}} & \multicolumn{2}{c}{\textbf{Zero-Shot}} \\
    \cmidrule{2-5}
    ~ & \textbf{24-layer} & \textbf{48-layer} & \textbf{24-layer} & \textbf{48-layer} \\
    \midrule
    2 & 28.01 / 24.20 & 29.51 / 25.58 & \;\;4.68 / 5.26\;\; & \;\;3.51 / 4.42\;\; \\
    3 & 28.16 / 24.46 & 29.38 / \textbf{25.73} & \;\;4.55 / 5.03\;\; & \;\;4.07 / 9.48\;\; \\
    4 & \textbf{28.21} / 24.72 & 29.28 / 25.68 & 16.19 / 17.25 & 10.34 / 6.19\;\;\\
    5 & 28.04 / \textbf{24.77} & \textbf{29.97} / 25.45 & \textbf{17.29} / \textbf{18.29} & 16.61 / 18.26 \\
    6 & 27.11 / 24.51 & 28.51 / 25.20 & 17.20 / 18.11 & \textbf{17.75} / \textbf{18.65} \\
    \bottomrule
    \end{tabular}
}
    \caption{
        BLEU scores of LCS with different $k$ on the noise and denoise OPUS-100.
        $k$ denotes the hyperparameter $k$ and 24/48-layer denotes the encoder's depth.
    }
\label{tab: lt_layers_deeper}
\end{table}

\begin{figure*}[t!]
\begin{center}
    \scalebox{0.875}{
    \subfigure[Many-to-One, T-Enc]{
        \centering
        \begin{minipage}[t]{0.3\textwidth}
        \includegraphics[width=\textwidth]{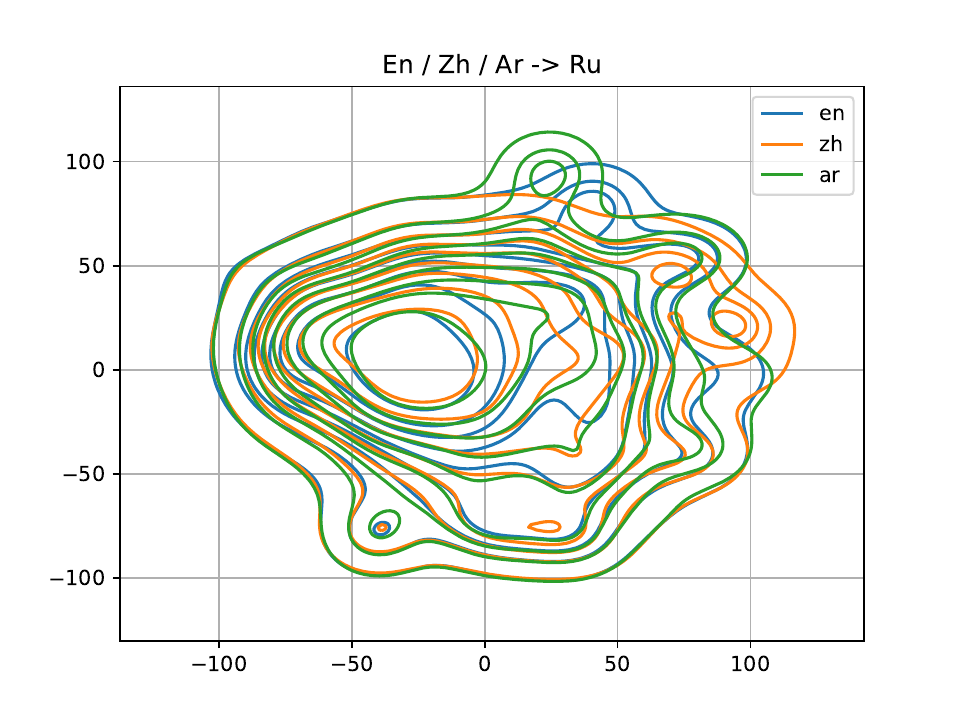}
        \end{minipage}
        \label{fig: tag1_m2o}
    }
    \subfigure[Many-to-One, S-Enc-T-Dec]{
        \centering
        \begin{minipage}[t]{0.3\textwidth}
        \includegraphics[width=\textwidth]{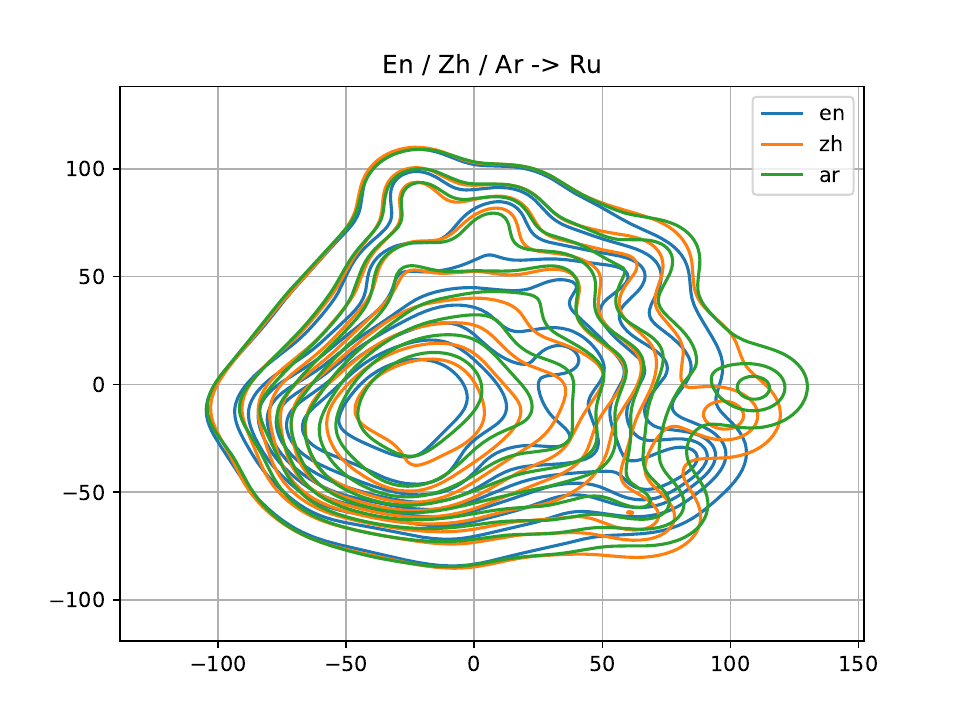}
        \end{minipage}
        \label{fig: tag2_m2o}
    }
    \subfigure[Many-to-One, LCS]{
        \centering
        \begin{minipage}[t]{0.3\textwidth}
        \includegraphics[width=\textwidth]{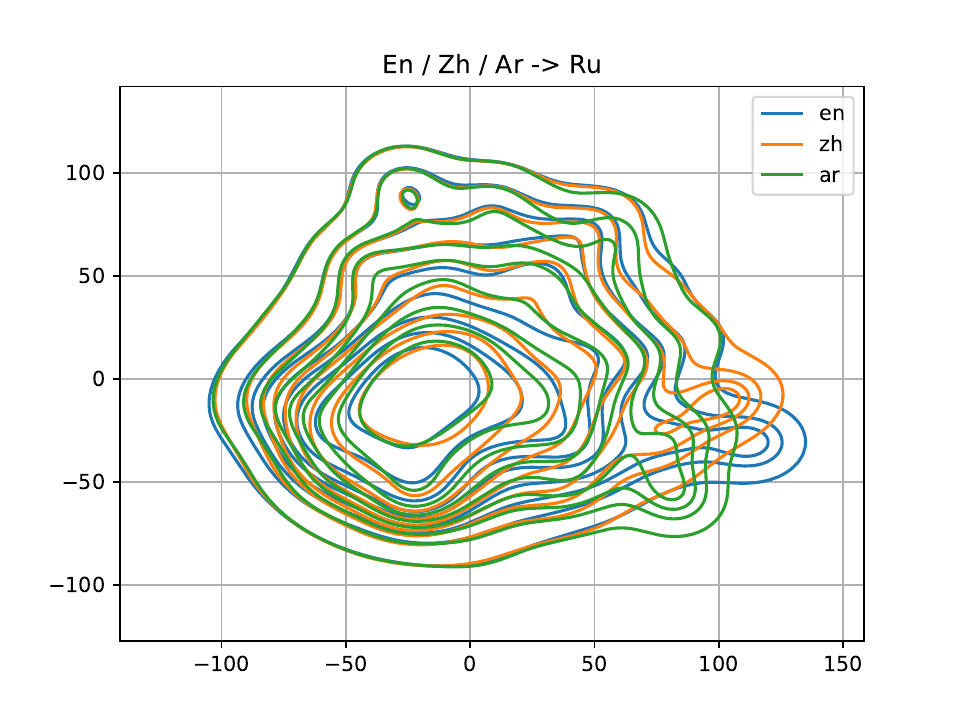}
        \end{minipage}
        \label{fig: lcs_m2o}
    }
    }
    \quad
    \scalebox{0.875}{
    \subfigure[One-to-Many, T-Enc]{
        \centering
        \begin{minipage}[t]{0.3\textwidth}
        \includegraphics[width=\textwidth]{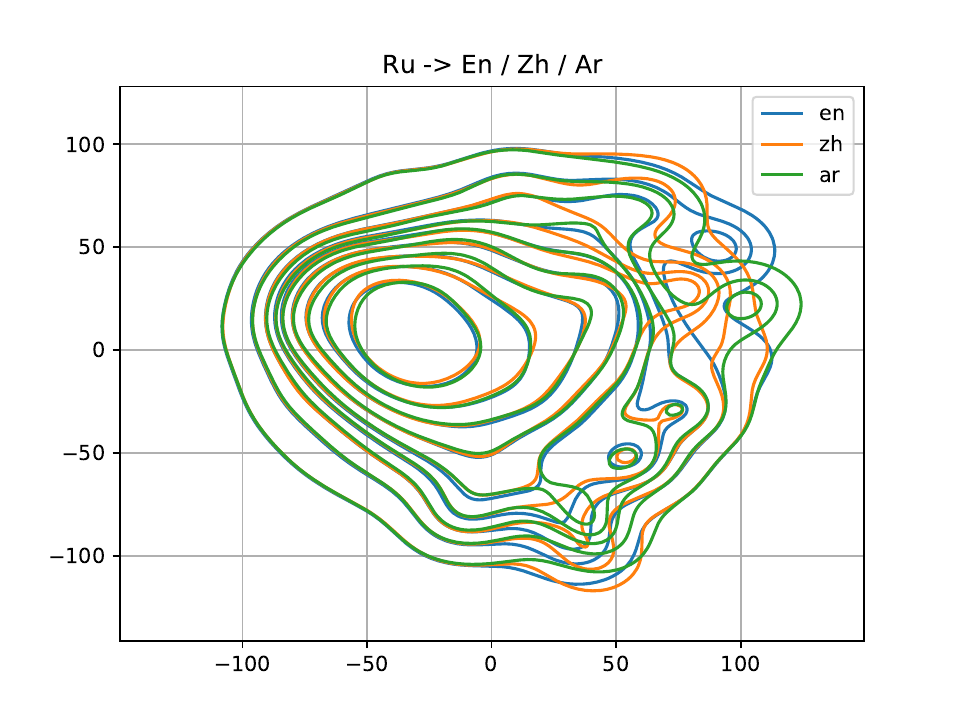}
        \end{minipage}
        \label{fig: tag1_o2m}
    }
    \subfigure[One-to-Many, S-Enc-T-Dec]{
        \centering
        \begin{minipage}[t]{0.3\textwidth}
        \includegraphics[width=\textwidth]{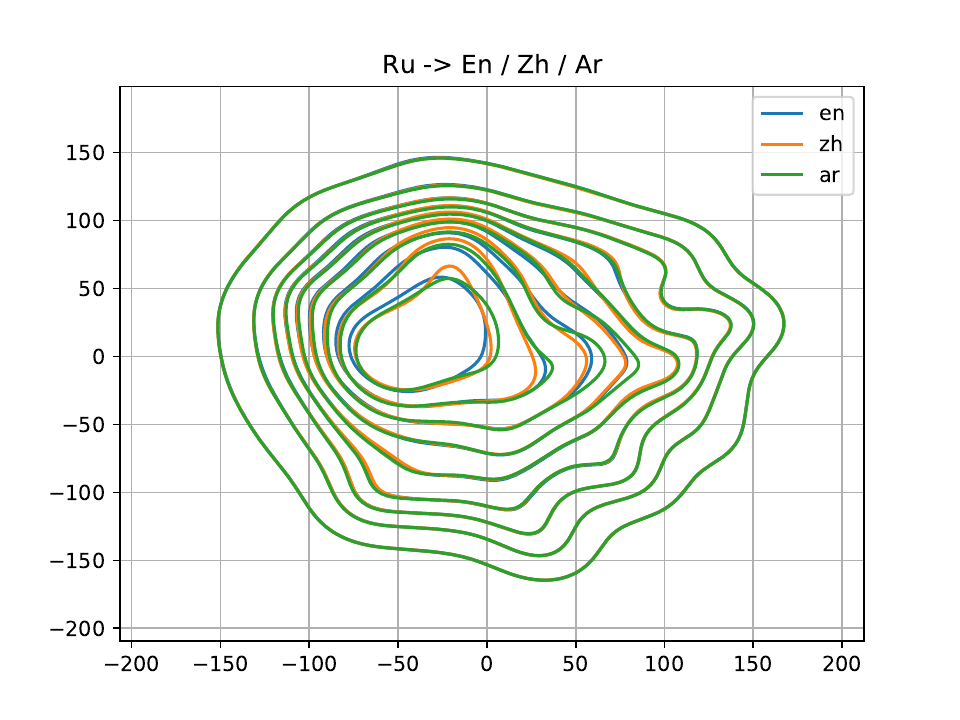}
        \end{minipage}
        \label{fig: tag2_o2m}
    }
    \subfigure[One-to-Many, LCS]{
        \centering
        \begin{minipage}[t]{0.3\textwidth}
        \includegraphics[width=\textwidth]{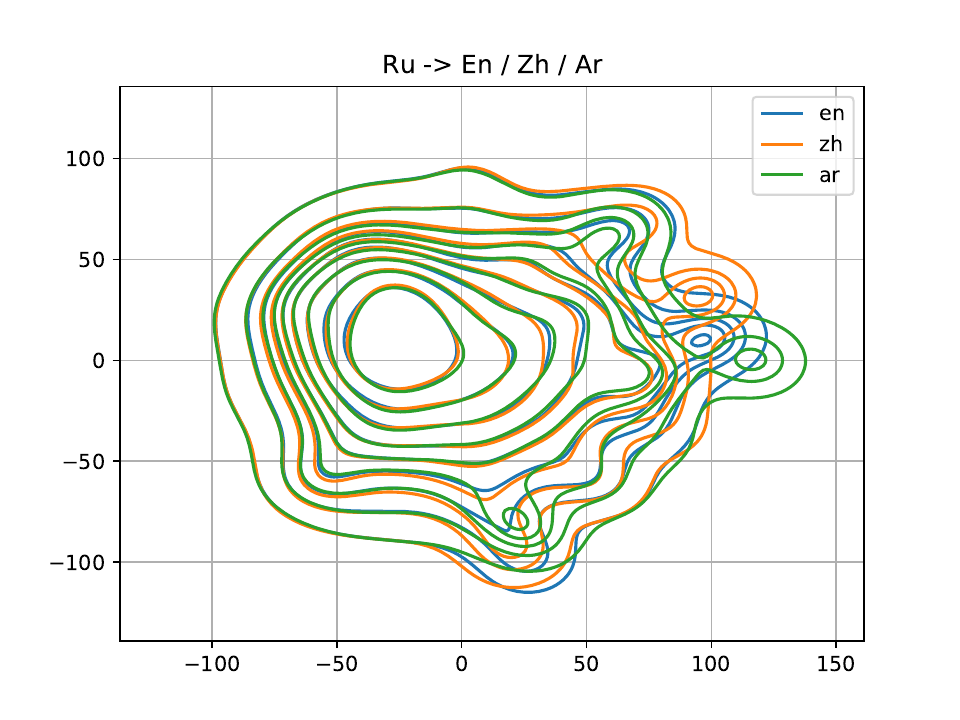}
        \end{minipage}
        \label{fig: lcs_o2m}
    }
    }
    \caption{
        The KDE of T-SNE reduced averaged encoder output in many-to-one and one-to-many directions.
    } 
    \label{fig: tsne}  
    \vspace{-2pt}
 \end{center} 
\end{figure*}

\begin{figure}[t!]
\begin{center}
    \scalebox{0.485}{
        \subfigure[] {
            \begin{minipage}[t]{\linewidth}
            \includegraphics[width=\textwidth]{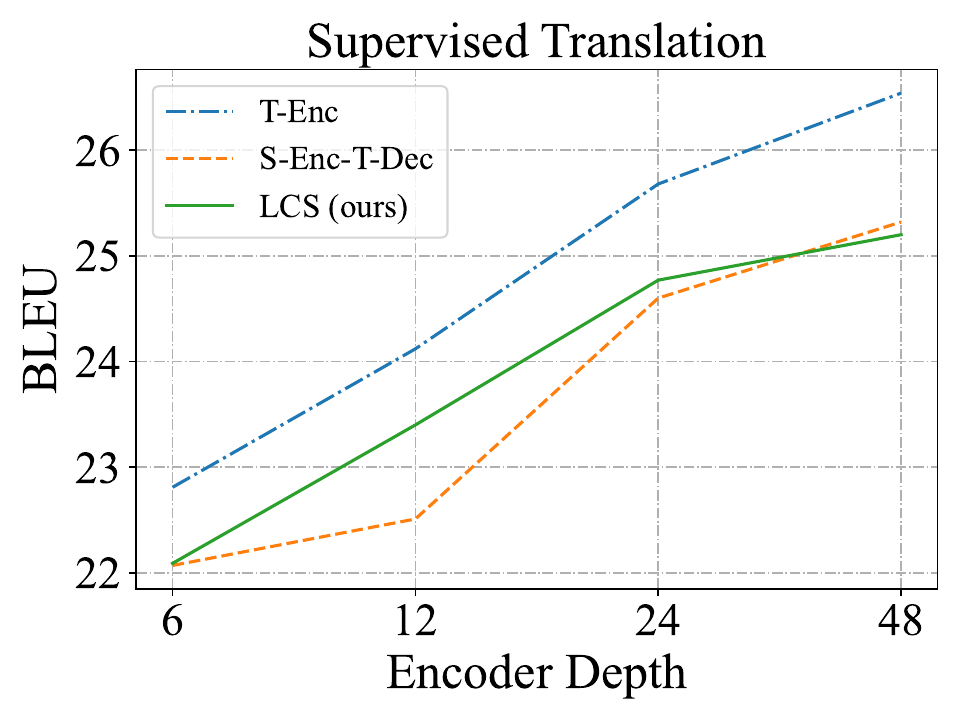}
            \end{minipage}
            \label{fig: supervised_trans_denoise}
        }
        \subfigure[] {
            \begin{minipage}[t]{\linewidth}
            \includegraphics[width=\textwidth]{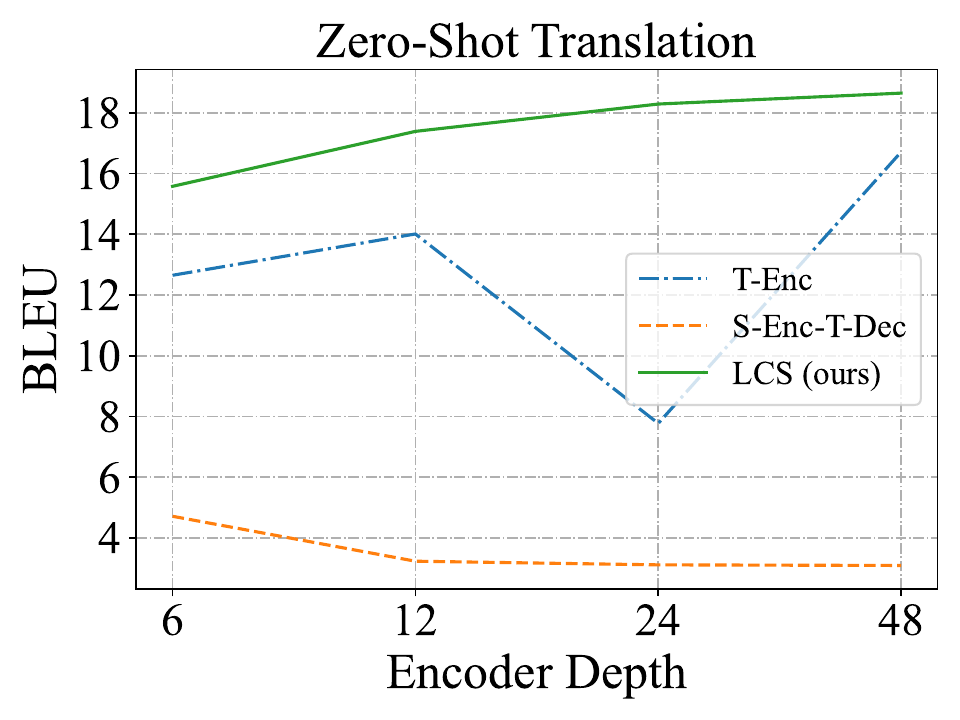}
            \end{minipage}
            \label{fig: zeroshot_trans_denoise}
        }
    }
    \caption{
    Translation performance of deeper encoder on the supervised and zero-shot test set of denoised OPUS-100.
    } 
    \label{fig: deep_trans_denoise}  
    \vspace{-6pt}
 \end{center} 
\end{figure}

\section{Selection of $k$ in deeper encoder} \label{sec: deeper_exps}
In this section, we supply the experiments of exploration on hyperparameter $k$ in 24- and 48-layer encoders on the noise and denoise OPUS-100 datasets.
We report the results in Tab.\ref{tab: lt_layers_deeper}.

As illustrated by Tab.\ref{tab: lt_layers_deeper}, the optimal selection of $k$ in the 24-layer encoder is 5 and 6 in the 48-layer encoder on zero-shot translation, while keeping similar performance on supervised translation.
The better range of selection of $k$ is the nearby integers of 15\% of the encoder depth.

Further, we also evaluate the performance of T-Enc, S-Enc-T-Dec, and LCS strategies on the denoised OPUS-100 dataset.
Compared to the noise version, although the T-Enc could roughly yield better performance of zero-shot translation with more encoder layers, LCS could perform better than it, demonstrating that LCS is effective in boosting the performance of zero-shot translation.

\section{T-SNE Visualization} \label{appendix: tsne}
To detect the language-specific of the encoder output representation, we sample 2000 samples from the MultiUN test set\footnote{Each sentence has the corresponding translation in every language.}, each with four languages, \emph{i.e.}, Arabic (Ar), English (En), Russian (Ru), and Chinese(Zh).
We retrieve the encoder output representation in T-Enc, S-Enc-T-Dec, and LCS strategies on many-to-one and one-to-many directions\footnote{Our models are trained in many-to-many direction}, and visualize them by T-SNE \citep{van2008visualizing}, following \citet{pan-etal-2021-contrastive}.
%

As shown in Figure \ref{fig: tsne}, in the many-to-one direction (Figure \ref{fig: tag1_m2o}, \ref{fig: tag2_m2o}, and \ref{fig: lcs_m2o}), the contour lines of three strategies tend to overlap, while the contour lines of T-Enc have the highest degree to overlap with each other, and the contour lines of S-Enc-T-Dec have the lowest degree.
The overlap results suggest that the language-specific of the encoder output representation is target-language-specific and the overlap ranking of the three strategy contour lines is consistent with the cosine similarity ranking in \ref{fig: variation}.
However, in the one-to-many direction of S-Enc-T-Dec (Figure \ref{fig: tag2_o2m}), the contour lines nearly perfectly overlap with each other, suggesting that the encoder output representation of S-Enc-T-Dec tends to be source-specific rather than target-language-specific with the indication from the source language tag in deep encoder layers.
In summary, the encoder output representation of the T-Enc and LCS strategy tends to be target-language-specific, while the representation of the S-Enc-T-Dec strategy tends to be source-language-specific.
\end{document}